\documentclass[conference]{IEEEtran}
\IEEEoverridecommandlockouts
\usepackage{cite}
\usepackage{amsmath,amssymb,amsfonts}
\usepackage{graphicx}
\usepackage{textcomp}
\usepackage{xcolor}
\usepackage{algorithm}%
\usepackage{algorithmicx}%
\usepackage{algpseudocode}
\usepackage{tabularx}
\usepackage{booktabs}
\usepackage{multirow}
\def\BibTeX{{\rm B\kern-.05em{\sc i\kern-.025em b}\kern-.08em
    T\kern-.1667em\lower.7ex\hbox{E}\kern-.125emX}}
    
\begin{document}

\title{Efficient and Robust Spiking Neural Networks for sEMG-Based Muscle Fatigue Detection}
\author{%
\IEEEauthorblockN{%
Kaiwen Tang\textsuperscript{*},
Jiaqi Dong\textsuperscript{*},
Zhanglu Yan\textsuperscript{\textdagger},
and Weng-Fai Wong
}
\IEEEauthorblockA{%
\textit{University of Singapore}\\
Singapore, Singapore
}
\thanks{%
\textsuperscript{*}Kaiwen Tang and Jiaqi Dong contributed equally to this work.
}%
\thanks{%
\textsuperscript{\textdagger}Corresponding author: Zhanglu Yan
(zhangluyan@comp.nus.edu.sg).
}%
}
\maketitle
\begin{abstract}
Detecting muscle fatigue via surface electromyography (sEMG) is essential for applications in sports, rehabilitation, and wearable health monitoring. Accurate and timely detection of fatigue is crucial for preventing injuries, optimizing physical performance, and ensuring user safety during prolonged activity. 
However, existing deep learning models are often unsuitable for this task due to their high computational cost and dependence on large-scale data. 
In this work, we propose an energy-efficient framework for muscle fatigue detection based on Spiking Neural Networks (SNNs), which exploit sparse, event-driven computation and temporal modeling. We further introduce a quantization-compatible training scheme (SDH) that combines multiple regularization terms to improve robustness under noisy conditions.
Evaluated on two public sEMG datasets against a broad set of baselines and under seven noise conditions including physically motivated perturbations, our quantized SNNs match or exceed strong baselines while remaining more stable under diverse noise and reducing estimated energy consumption by up to 201.77×. These results demonstrate the framework's strong potential for real-time deployment in low-power wearable systems.

\end{abstract}


\section{Introduction}\label{main}

Muscle fatigue detection is essential across domains such as sports, rehabilitation, and occupational safety~\cite{wan2017muscle, guan2021sports, bangaru2022automated}. In competitive sports, timely identification of fatigue allows coaches to optimize training loads and prevent injuries; in labor-intensive industries, it helps managers mitigate safety risks by monitoring workers' physical conditions. Surface electromyography (sEMG), a non-invasive technique for recording muscle electrical activity~\cite{al2011review}, provides a promising signal source for fatigue assessment. However, sEMG signals are often noisy, subject-dependent, and exhibit complex temporal dynamics, posing challenges for reliable real-time analysis in practical environments, particularly when the system is expected to run continuously on power-constrained, wearable devices.

In real-world applications such as sports training or workplace monitoring, muscle fatigue detection must operate reliably on lightweight wearable devices under unconstrained, noisy conditions~\cite{moniri2020real, seshadri2019wearable}. To preserve user privacy and enable low-latency feedback, models must process data locally rather than relying on cloud-based computation~\cite{zhang2015real}. This imposes three core requirements on practical models: (1) high predictive accuracy under limited training data; (2) low computational overhead suitable for edge devices; and (3) strong robustness to signal noise induced by motion artifacts or environmental factors.

Prior work in fatigue detection primarily leverages either conventional machine learning (ML) models or deep neural networks. 
Conventional ML methods such as SVM, k-NN, or ensemble methods~\cite{puce2021surface, papakostas2019physical, chen2021psychophysiological} often have limited ability to capture the temporal dynamics inherent in fatigue progression. 
While deep models such as LSTM~\cite{wang2021muscle} and attention-based methods~\cite{zhang2021mffnet} can improve predictive performance, they usually come with higher latency and energy cost, which limits their use on resource-constrained wearable platforms. Moreover, due to the non-invasive nature of sEMG acquisition, the signals remain highly susceptible to noise~\cite{rozaqi2019design}, and few existing methods explicitly address robustness to motion-induced artifacts~\cite{corvini2022estimation, anwer2024evaluation}. As a result, reliable real-time on-device deployment remains challenging.

Spiking Neural Networks (SNNs) offer a promising solution by mimicking the event-driven communication of biological neurons. Their sparse spike-based computation enables low-power, temporally-aware signal processing~\cite{eshraghian2023training, tang2024onespike}. SNNs have achieved strong results in bio-signal applications such as EEG-based emotion recognition~\cite{xu2024eescn} and ECG-based arrhythmia detection~\cite{fan2024ultra}, benefiting from quantization-aware training~\cite{chu2023energy} and hybrid-response architectures~\cite{li2024hr}. Recent studies demonstrate their applicability to sEMG~\cite{xu2023novel, guo2024spgesture}, achieving real-time inference with reduced energy consumption. However, prior SNN models for sEMG either focus on gesture classification or require specialized encoding schemes, and few are tailored to the noise-prone, data-limited setting of muscle fatigue detection.

In this work, we propose a compact, quantized SNN framework for real-time, on-device muscle fatigue detection. By first training a surrogate ANN and converting it into an SNN via a structure-preserving pipeline~\cite{yan2023cq}, our method retains the predictive performance of the ANN while benefiting from the event-driven nature of SNNs for significant energy efficiency. To further address noise and signal variability in sEMG data, we introduce a robust training scheme named SDH, which integrates safe haven activation quantization (SHAQ)\cite{yan2024improving}, defensive quantization (DQ)\cite{lin2019defensive}, and Huber loss. We apply 3-bit and 4-bit weight quantization to further compress the model and improve deployability on low-power platforms. 

We conduct experiments on two public sEMG datasets, SUE and SPE, against a broad set of baselines. 
Our weight quantized spiking models, denoted as QSN3 and QSN4, achieve strong performance on both datasets. In particular, the spike-based models reach F1 scores of 87.93\% on SUE and 90.14\% on SPE, compared with 87.36\% and 87.00\% from the best non-spiking neural baselines under QAF, respectively.
For robustness evaluation, we test seven noise types. The results show that the proposed spike-based models are more stable under diverse noisy conditions. We further study energy efficiency. The estimated energy consumption is reduced by up to 201.77$\times$, supporting the suitability of the proposed models for resource-constrained wearable deployment. We also compare the energy cost of the competing methods and estimate battery life under a smartwatch-scale energy budget.
To the best of our knowledge, our approach is the first to offer an energy-efficient, edge-device-compatible SNN solution for sEMG-based muscle fatigue detection with enhanced noise resilience. 

Our contributions are summarized as follows: 

\begin{itemize}

\item We establish spiking neural networks as a promising paradigm for muscle fatigue detection and propose a series of spike-based models. Our approach effectively reduces estimated energy consumption by up to 201.77$\times$ compared with the full-precision ANN baseline.

\item We introduce a novel training scheme called SDH, which significantly boosts the model’s resilience under both synthetic and physically motivated perturbations.

\item Experiments on the public SUE and SPE datasets show that the proposed models achieve competitive F1 scores, remain robust under diverse noise conditions, and substantially reduce estimated energy consumption. 

\end{itemize}

\begin{figure*}[h]
    \centering
    \includegraphics[width=0.8\textwidth]{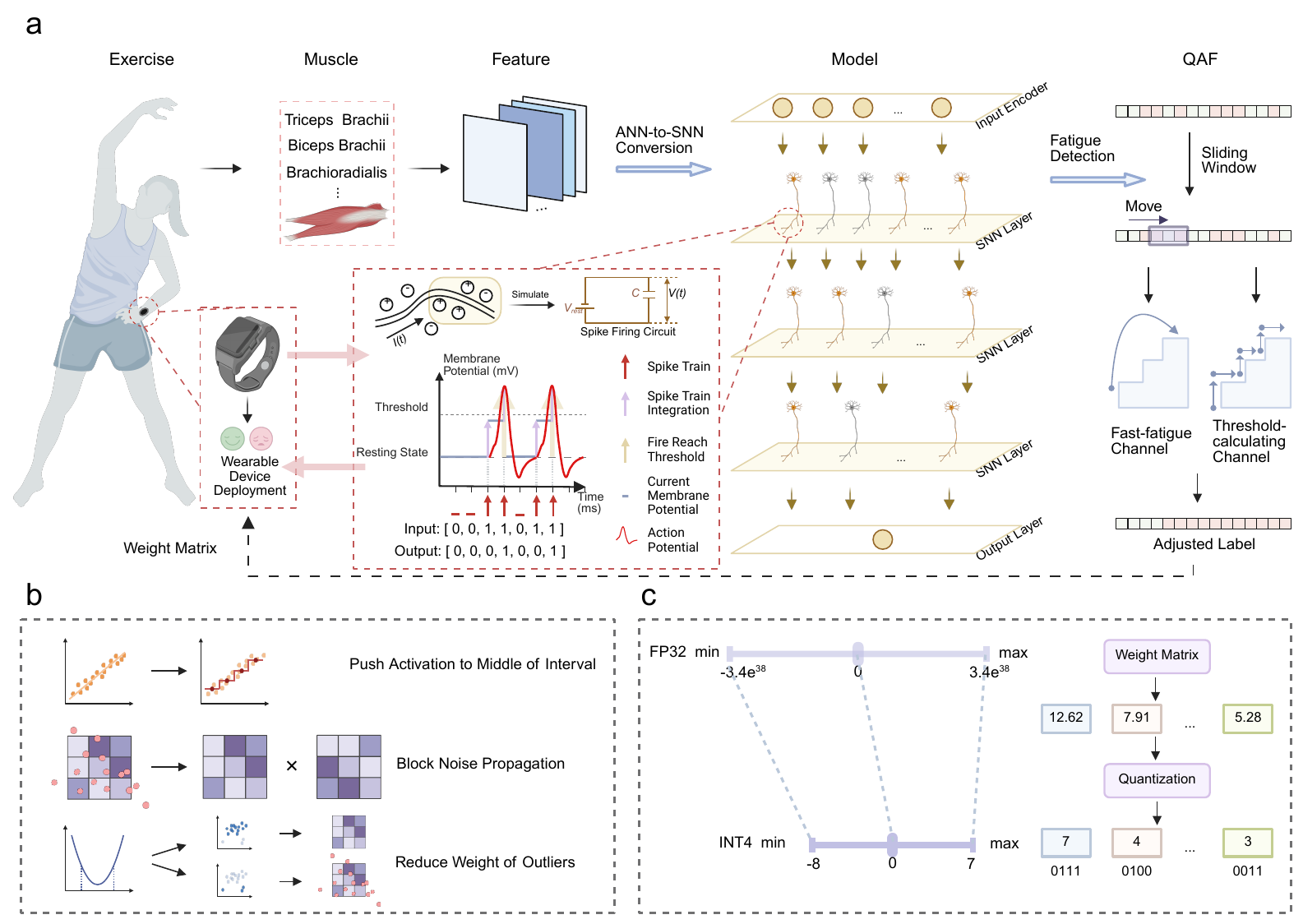}
    \caption{Overview of the proposed muscle fatigue detection framework. \textbf{a.} Personalized SNN pipeline with feature extraction, spike encoding, SNN inference, and QAF post-processing.
    \textbf{b.} The three components of our robustness enhancement method, SDH. \textbf{c.} Low-bit weight quantization from FP32 to signed integers.
    }
    \label{fig:main}
\end{figure*}

\section{Preliminary}
\label{Preliminary}
Spike neurons mimic a biological neuron's behavior by integrating incoming signals and emitting a discrete spike. 
The IF model is a simplified mathematical model of a spiking neuron that captures the firing behavior of neurons in the brain, which is the most widely used for generating spike trains~\cite{bu2023optimal}. It updates a neuron's membrane potential and generates a spike when a threshold is reached. First, the membrane potential \(V_i^l(t)\) is updated by accumulating the neuron's input activation \(A_i^l\) with its previous potential \(V_i^l(t-1)\):
\begin{equation}
V_i^l(t) = V_i^l(t-1) + A_i^l.
\end{equation}

When \(V\) exceeds a specified threshold $\theta$, the neuron generates a spike. The output spike \(s_j^{l+1}(t)\) for the neuron in the subsequent layer is defined as:
\begin{equation}
s_j^{l+1}(t) =
\begin{cases}
1, & \text{if } V_i^l(t) \geq \theta, \\
0, & \text{otherwise},
\end{cases}
\end{equation}
where \(\theta\) is the firing threshold. When the neuron generates spikes, the membrane potential is reduced by \(\theta\) (emulating a reset); otherwise, it remains unchanged:
\begin{equation}
V_i^l(t) =
\begin{cases}
V_i^l(t) - \theta, & \text{if } V_i^l(t) \geq \theta, \\
V_i^l(t), & \text{otherwise}.
\end{cases}
\end{equation}


\section{Methods}
\label{Methods}
In this section, we begin by presenting an overview of our framework in Section~\ref{Methods1}. Next, we detail the construction of our SNN in Section~\ref{SNN_construction}. In Section~\ref{robustness} and Section~\ref{quantization}, we introduce our proposed method for enhancing robustness and our quantization process. Finally, we describe QAF, our post-processing method in Section~\ref{Post-processing}.

\subsection{Framework Overview}
\label{Methods1}
In our work, the pipeline consists of four main stages: feature extraction, SNN construction, model quantization, and post-processing as shown in Figure~\ref{fig:main}. For feature extraction, we follow the same assumptions as previous studies, extracting short-term and mid-term features from sEMG signals. Within the SNN, each spiking neuron adopts an averaged IF model to encode inputs into spike trains, which are passed through a sigmoid activation function to yield a probability score. Additionally, we incorporate a noise-resistant training method as in Figure~\ref{fig:main}b and perform post-training weight quantization as in Figure~\ref{fig:main}c. Finally, we introduce a post-processing method named {\em quick adapt filter} (QAF). Based on the post-processing in previous work~\cite{papakostas2019physical}, QAF dynamically adjusts channels and thresholds to enhance detection accuracy and minimize latency. 

\subsection{SNN construction}
\label{SNN_construction}

Before input to each layer, all the data are encoded into spike trains consisting of 0s and 1s. In this work, we adopted a variant of the IF model, namely {\em average spike generation} (ASG)~\cite{yan2022low}. 
In ASG, there is an additional step before the IF model, which is to take the average of all the membrane potentials and use this average value as the new input to generate spike trains. 
In the constructed SNN model, the encoded spike train serves as the input and propagates through the network layer by layer. At each layer, the input spike train interacts with the weights, resulting in the computation of the membrane potential ($V$) as shown in Figure~\ref{fig:main}a. With the ASG algorithm shown above, it is further encoded into the output spike trains. These output spike trains are then transmitted to the subsequent layer as input. 

Training SNNs directly is challenging due to their non-differentiable nature. To address this, we adopt an ANN-to-SNN conversion method. 
This method first trains a traditional ANN and then converts it into an SNN with the same structure for inference.
During training, we use clamp and quantization functions as activation function to optimize the weight transfer and reduce computational complexity~\cite{yan2023cq}. Specifically, the activation functions can be represented as:
\begin{equation}
    CQ(x) = clamp(\frac{\lfloor x \cdot T \rfloor}{T}, 0, \theta)
\end{equation}
Where $T$ is the length of the spike train, also known as the time window size while $\theta$ is the spike threshold.
The architecture is a four-layer fully connected network (132-128-32-1), totaling 21.19K parameters, which is trained with an Adam optimizer and binary cross-entropy as a loss. 

\subsection{Robustness enhancement}
\label{robustness}
Wearable sEMG signals are often corrupted by electrode-skin impedance variations and motion-induced artifacts, which can degrade feature quality and model reliability~\cite{boyer2023reducing}. 
To improve robustness under such distortions, we introduce SDH, a noise-robust training objective designed to stabilize fatigue detection in unconstrained wearable settings.

Previously, the SHAQ~\cite{yan2024improving} and DQ~\cite{lin2019defensive} methods introduced penalty terms in the loss function to improve noise resistance for other tasks. 
Specifically, SHAQ pushes activation values towards more noise-resilient ranges within the quantization interval, while DQ minimizes the absolute values of weights to limit the propagation of noise across layers. 
However, neither method achieved optimal performance in muscle fatigue detection.
To address this, we incorporate Huber loss into the training objective. The Huber loss behaves quadratically for small residuals and linearly for large ones, which makes it particularly effective at suppressing the influence of transient spikes or outliers. 
These terms target complementary sensitivities: $L_{act}$ stabilizes quantized activations, $L_{orth}$ limits cross-layer perturbation propagation, and $L_{huber}$ down-weights abrupt outliers.
We combine them in an additive form so that each component targets a different source of sensitivity while keeping the training objective simple and differentiable. Thus, the loss function of SDH can be written as

\begin{equation}
    \mathcal{L} = \mathcal{L_{\text{BCE}}} + \lambda_1 \cdot \mathcal{L_{\text{act}}} + \lambda_2 \cdot \mathcal{L_{\text{orth}}} + \lambda_3 \cdot \mathcal{L_{\text{huber}}}
\end{equation}

In this composite loss function, \(\mathcal{L}_{\text{BCE}}\) is the binary cross-entropy term used for classification.
The term $\mathcal{L_{\text{act}}}$ is the activation regularization term derived from SHAQ, defined as: $L_{\text{act}} = \sum \left| \textbf{x} - \text{seq}(\textbf{x}) \right|$, where \(\mathrm{seq}(\mathbf{x})\) is the midpoint of the quantization interval to which the activation is mapped. By moving activations closer to the middle of their quantization intervals, this term makes the quantized representation less sensitive to small input perturbations.

The term \(\mathcal{L}_{\text{orth}}\) is the weight regularization term derived from DQ, defined as $L_{\text{orth}} = \sum \left\| \textbf{W}^T \textbf{W} \right\|_F^2$. This formulation follows the defensive quantization strategy in~\cite{lin2019defensive}. Unlike the standard orthogonality constraint \(\|\mathbf{W}^{T}\mathbf{W}-\mathbf{I}\|_{F}^{2}\), the objective here is not to enforce strict orthonormality. Instead, it penalizes large responses in \(\mathbf{W}^{T}\mathbf{W}\), thereby suppressing excessively large weights and reducing correlations among weight vectors. This helps limit noise propagation across layers. At the same time, by discouraging highly correlated weight directions, the term encourages more independent feature representations, which can improve generalization and reduce overfitting.

Finally, the $L_{\text{huber}}$ term represents the Huber loss, defined as:
\begin{equation}
    L_{\text{huber}}(x) =
\begin{cases}
\frac{1}{2} x^2 & \text{if } |x| \leq \delta \\
\delta (|x| - \frac{1}{2}\delta) & \text{otherwise}
\end{cases}
\end{equation}
Here, \(\delta\) is a hyperparameter that controls the transition point of the loss function. By reducing sensitivity to abrupt fluctuations caused by noise, the Huber term allows the model to focus more on the overall trend of the fatigue signal rather than isolated outliers.

The parameters \(\lambda_1\), \(\lambda_2\), \(\lambda_3\), and \(\delta\) control the contributions of the three regularization terms and the transition point of the Huber loss. In practice, these coefficients are determined by grid search on the validation set under uniform noise only. The selected coefficients are then fixed for the remaining noise types. This protocol avoids noise-specific tuning and provides a stricter test of whether the learned robustness generalizes beyond the perturbation used for coefficient selection.

\subsection{Model quantization}
\label{quantization}
Quantization is essential for deploying SNNs on edge devices, as it reduces memory usage and accelerates computation by lowering the bit-width of weights and biases. We adopt post-training quantization (PTQ) for its simplicity and compatibility with pre-trained models, avoiding the need for retraining. The inherent sparsity of spike trains further limits the impact of quantization noise on model performance. While discretization introduces errors, selecting an appropriate bit-width $n_{\mathrm{bit}}$ allows a balance between accuracy and efficiency.
In our experiments, a bit-width of $n_{\text{bit}} = 4$ preserves high accuracy with reduced memory cost, while $n_{\text{bit}} = 3$ remains effective in most cases, with occasional instability.









\subsection{Post-processing}
\label{Post-processing}


Post-processing is applied to suppress short-term classification noise and improve fatigue detection stability. We first use a median filter with \( K=3 \) to smooth transient fluctuations, followed by a window-based adjustment strategy: classification results are grouped into sliding windows of size \( M=3 \), and a fatigue state is triggered if the number of ``fatigue'' labels exceeds a threshold over the preceding \( N \) windows. This method performs well under gradual fatigue accumulation but exhibits delayed response when fatigue develops rapidly.

To address this, we introduce a ``fast channel'' mechanism that allows immediate state transitions when a strong fatigue signal is detected within a single window. Furthermore, we replace the fixed threshold with an adaptive one computed via an exponentially weighted moving average (EMA)~\cite{li2022application}, allowing the system to track recent trends more responsively. The threshold at time $t$ is updated as:
\begin{equation}
T_t = \alpha \cdot r_t + (1 - \alpha) \cdot T_{t-1}
\end{equation}
where $r_t$ is the proportion of “fatigue” labels in the current window and \(\alpha = 0.2\) is the smoothing factor. This formulation improves the system’s sensitivity to early signs of fatigue while maintaining stability. The full procedure, named QAF, is outlined in Algorithm~\ref{alg3}.

\begin{algorithm}
\caption{Fatigue Detection Post-Processing}\label{alg3}
\begin{algorithmic}[1]
\Require
Fatigue predictions $P$, window size $w$, step size $\Delta$, base threshold $\theta_b$, smoothing factor $\alpha$, fast threshold $\theta_f$
\Ensure Post-processed predictions $P'$

\State Initialize: threshold $T \gets \theta_b$, start index $s \gets 0$
\State Previous windows $p_1, p_2 \gets \text{nil}$

\While{$s + w \leq \text{len}(P)$}
    \State Current window $W \gets P[s:s+w]$, ratio $r \gets \frac{\text{sum}(W)}{w}$

    \If{$r \geq \theta_f$}
        \State $P' \gets \{ s \times [0], (\text{len}(P) - s) \times [1] \}$
        \State \textbf{break}
    \EndIf

    \State Update threshold: $T \gets \alpha \cdot r + (1 - \alpha) \cdot T$

    \If{$p_1 \neq \text{nil} \land p_2 \neq \text{nil}$}
        \If{$r \geq T \land \frac{\text{sum}(p_1)}{w} \geq T \land \frac{\text{sum}(p_2)}{w} \geq T$}
            \State $P' \gets \{ s \times [0], (\text{len}(P) - s) \times [1] \}$
            \State \textbf{break}
        \EndIf
    \EndIf

    \State Update previous windows: $p_2 \gets p_1$, $p_1 \gets W$
    \State Move window: $s \gets s + \Delta$
\EndWhile

\If{$s + w > \text{len}(P)$}
    \State $P' \gets \{\text{len}(P) \times [0]\}$
\EndIf
\end{algorithmic}
\end{algorithm}

\subsection{Architecture Design}
\label{sec: arc}
\begin{figure}[h]
    \centering
    \includegraphics[width=0.5\textwidth]{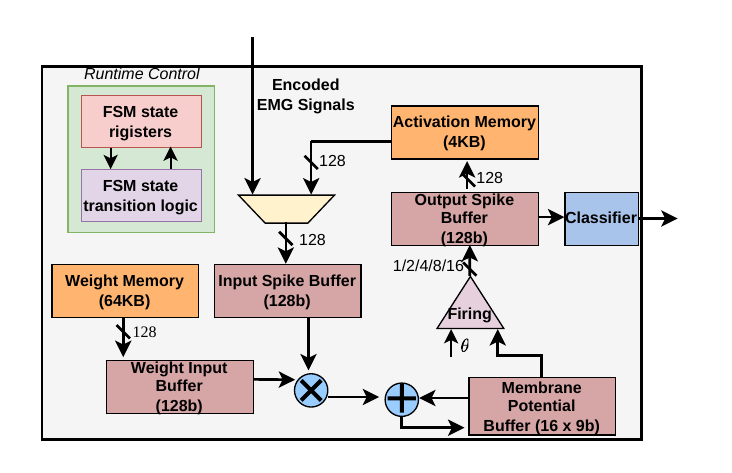}
    \caption[]{CU design and associated scheduler.}
    \label{fig:CU-design}
\end{figure}

Figure~\ref{fig:CU-design} shows the CU datapath and scheduler in our SNN.
To minimize power and area, our SNN setting uses a single CU for all computations; for the EMG classification workloads, this CU sustains the required output rate. The CU provides dedicated datapaths for accumulation, bias addition, spike generation, and classification. A scheduler FSM selects the active datapath, generates memory addresses, and controls SRAM reads/writes.

The FSM has five stages. In \textit{ACC}, weights and activations are streamed through input buffers refilled from SRAM. After all source neurons for an output neuron are processed, \textit{BIAS} adds the bias in one cycle. In \textit{SPIKE}, the membrane potential is updated, compared with the threshold, and written bit-wise to the output buffer; when a spike is generated, the threshold is subtracted from the membrane potential. In \textit{SAVE}, the output spike vector is written to double-buffered activation memory to avoid overwriting the previous layer. After the final layer, \textit{CLASSIFY} outputs the EMG classification results.


For the Memory Organization, 
weighted-sum computation dominates memory traffic because each source--target neuron pair requires the corresponding activation and weight. Biases, thresholds, and output activations are accessed less frequently: once per neuron, once per layer, and once per neuron, respectively. Our SNN setting therefore separates immutable parameters from activations. Weights, biases, and thresholds are stored in read-only SRAM with a 128-bit port, while activation SRAM is also buffered at 128 bits to match the memory width. This amortizes per-access energy, as narrower ports incur higher marginal energy per bit. Activation input/output buffers expose configurable 1--16-bit ports to support different activation precisions.
To improve locality, the CU includes a $16 \times 9$-bit membrane-potential buffer, where $16$ supports the selected deployment setting with $T=15$, and $9$ is the accumulator precision for 4-bit accumulation.
After all neurons from preceding layers are accumulated, the buffer is forwarded to the spike-generation datapath. This local buffering avoids repeated activation-SRAM accesses and enables efficient event-driven accumulation.

\section{Experiment}
\label{experiment}

\subsection{Datasets}
\label{datasets}
We conduct experiments on two publicly available sEMG datasets: SUE\cite{papakostas2019physical} and SPE\cite{lim2024assessment}. SUE includes data from 10 participants performing three upper-limb exercises: shoulder flexion, shoulder abduction, and elbow extension, using a robotic arm that provides resistance feedback. sEMG signals were recorded at 1926Hz from the deltoid, triceps, and related muscle groups. In each trial, participants moved to a target position and maintained it until self-reported fatigue, confirmed by research staff. Each exercise was repeated three times with rest intervals, resulting in 90 labeled recordings.

SPE contains data from 30 participants, with sEMG signals sampled at 2148Hz using Delsys Trigno Avanti sensors placed on nine upper-body muscles. A physiotherapist performed palpation assessments during the tasks, and participants later reported their top three fatigued muscles. We adopt the physiotherapist’s tenderness score as ground truth. Based on the 4-point scale~\cite{bendtsen1995pressure}, scores of 0–2 indicate non-fatigue, while 3 denotes visible pain and is labeled as “fatigued.” The dataset comprises 270 labeled recordings.

\subsection{EMG Signal Preprocessing and Feature Extraction}
\label{Preprocessing}
To suppress noise, raw sEMG signals are first smoothed with a median filter. The signals are then divided into non-overlapping short-term windows of duration \( T_s \) (0.25s for SUE, 7.5s for SPE). 
From each window, \( F'_s = 11 \) fatigue-related features are extracted~\cite{sun2022application,zhang2021mffnet}, including frequency-domain metrics (median/mean frequency), time-domain measures (ZCR, WAMP), spectral flux, energy entropy, and basic statistics (mean, max, min, FFT descriptors). These features are sensitive to physiological manifestations of fatigue. For example, reduced conduction velocity and prolonged motor unit action potentials shift the spectrum toward lower frequencies, decreasing median/mean frequency~\cite{xu2021advances}. Similarly, spectral flux, ZCR, WAMP, and entropy typically decline as the signal stabilizes and energy concentrates~\cite{ma2024laguerre,sonmezocak2021machine}.  We further compute delta features between adjacent windows, yielding $F_s=22$ short-term features in total.

To capture longer-term trends, short-term features are aggregated into overlapping mid-term windows of duration \( T_m \), with step size \( S_m = T_m / 2 \), yielding a 50\% overlap. Specifically, \( T_m = 2\,\text{s} \) and \( S_m = 1\,\text{s} \) for SUE, while \( T_m = 60\,\text{s} \) and \( S_m = 30\,\text{s} \) for SPE. From each mid-term window, six statistical descriptors are computed for each short-term feature:
\begin{equation}
\{ \mu, \sigma, \max, \min, \mu_{L}, \mu_{U} \}
\end{equation}
where \( \mu_L \) and \( \mu_U \) denote the means of the lower and upper third of values, respectively.
With $F_s=22$ total short-term features, the final mid-term feature vector has a dimension of $F=6\times F_s=132$.

Each mid-term feature vector is treated as an independent input to the SNN, generating one batch per second. Unlike prior studies that fed entire recordings as a single batch, this per-window approach reduces overfitting by focusing on intrinsic temporal patterns. 

\subsection{Results}
\label{Results}
\subsubsection{Model performance}
\label{performance}

We first assess all models on clean datasets for evaluation. Following prior work, we implemented traditional machine learning baselines, including Gradient Boosting, SVM with RBF kernel, Random Forest, Extra Trees, and KNN based on the models in~\cite{papakostas2019physical}, and further included MLP, 1D-CNN, LSTM, Transformer, and SpikeFormer as baselines. Table~\ref{tab:performance} reports the average F1 scores under three settings, namely the raw model output, the post-processing method from previous work~\cite{papakostas2019physical}, and our proposed QAF.


As shown in Table~\ref{tab:performance}, the proposed spike-based models achieve the best or near-best QAF F1 scores on both datasets. Compared with the ANN baselines, SNN improves F1 from 85.30\% to 87.93\% on SUE and from 87.00\% to 90.14\% on SPE, while QSN4 and QSN3 largely preserve performance after low-bit quantization.

\begin{table*}[h]
    \centering
    \caption{Performance comparison on the SUE and SPE datasets in terms of F1 score (\%). \textbf{Raw} denotes direct model output, \textbf{Post} denotes the post-processing method in~\cite{papakostas2019physical}, and \textbf{QAF} denotes the proposed post-processing method in Section~\ref{Post-processing}. Results are averaged over all user-specific models, and the MLP baseline together with the proposed models are further averaged over three fixed random seeds.}
    \label{tab:performance}
    \begin{tabularx}{\textwidth}{l|l|>{\centering\arraybackslash}X>{\centering\arraybackslash}X>{\centering\arraybackslash}X|>{\centering\arraybackslash}X>{\centering\arraybackslash}X>{\centering\arraybackslash}X}
        \hline
        \multirow{2}{*}{} & \multirow{2}{*}{Method} & \multicolumn{3}{c|}{SUE} & \multicolumn{3}{c}{SPE} \\
        \cline{3-5} \cline{6-8}
         & & Raw & Post & QAF & Raw & Post & QAF \\
        \hline

        \multirow{6}{*}{Classical ML} 
        & Gradient Boosting & 71.86 & 73.17 & 72.27 & 79.02 & 79.18 & 84.86 \\
        & SVM-RBF           & 72.07 & 79.65 & 79.02 & 82.42 & 77.23 & 86.55 \\
        & Random Forest     & 73.37 & 71.78 & 73.04 & 83.76 & \textbf{81.21} & 87.60 \\
        & Extra Trees       & 74.66 & 72.34 & 73.59 & 81.58 & 76.74 & 86.65 \\
        & KNN               & 72.38 & 81.16 & 82.69 & 79.29 & 79.28 & 85.36 \\
        \hline 
        \multirow{4}{*}{Neural Models}
        & MLP               & 73.74 & 87.14 & 85.30 & \textbf{83.78} & 77.35 & 87.00 \\
        & CNN1D             & 73.12 & 77.31 & 79.43 & 62.70 & 76.67 & 74.17 \\
        & LSTM              & \textbf{80.82} & 75.58 & 87.36 & 70.65 & 75.81 & 75.19 \\
        & Transformer       & 79.84 & 79.01 & 86.11 & 69.00 & 80.49 & 81.43 \\
        & SpikeFormer       & 80.69 & 78.18 & 87.81 & 70.73 & 80.73   & 80.94 \\
        \hline

        \multirow{3}{*}{Ours} 
        & SNN               & 74.28 & \textbf{88.62} & \textbf{87.93} & 83.15 & 79.11 & \textbf{90.14} \\
        & QSN4              & 74.20 & 86.19 & 87.84 & 83.33 & 78.98 & 90.00 \\
        & QSN3              & 70.98 & 86.24 & 87.39 & 83.01 & 78.58 & 89.13 \\
        \hline
    \end{tabularx}
\end{table*}


\begin{table}[t]
\centering
\caption{\textbf{Noise robustness} on SPE and SUE in terms of F1 score (\%). Bold marks the best result per noise and dataset.}
\label{tab:noise}
\scriptsize
\setlength{\tabcolsep}{2.2pt}
\renewcommand{\arraystretch}{0.92}
\begin{tabular}{@{}llccccc@{}}
\toprule
Data & Noise & MLP & SpikeFormer & SNN & QSN4 & QSN3 \\
\midrule
SPE & Gauss   & 87.93 & 89.17 & 90.30 & 90.21 & \textbf{90.43} \\
    & Poisson & 88.11 & 89.01 & 88.88 & \textbf{89.91} & 88.03 \\
    & Uniform & 87.83 & 89.72 & \textbf{90.57} & 90.43 & 89.81 \\
    & Multip. & 88.02 & 90.22 & 90.33 & \textbf{90.51} & 89.44 \\
    & Motion  & 87.88 & 90.30 & 90.77 & \textbf{91.01} & 90.65 \\
    & Power   & 88.16 & 89.52 & 90.26 & \textbf{90.76} & 89.20 \\
    & Drift   & 87.83 & 90.55 & 90.62 & \textbf{91.26} & 89.82 \\
\midrule
SUE & Gauss   & 86.44 & 83.20 & \textbf{88.18} & 87.07 & 87.05 \\
    & Poisson & 86.74 & 80.51 & 87.57 & 87.10 & \textbf{88.25} \\
    & Uniform & 85.10 & 83.06 & 89.63 & \textbf{89.73} & 89.38 \\
    & Multip. & 85.23 & 81.66 & 89.39 & \textbf{89.80} & 88.79 \\
    & Motion  & 84.43 & 85.17 & 89.13 & 88.46 & \textbf{89.38} \\
    & Power   & 85.36 & 80.06 & \textbf{89.87} & 89.07 & 88.17 \\
    & Drift   & 84.37 & 84.61 & \textbf{90.48} & 90.09 & 88.14 \\
\bottomrule
\end{tabular}
\vspace{-2mm}
\end{table}

\subsubsection{Noise robustness}
\label{Robustness_result}
We evaluate robustness under test-time signal corruption; no noise is injected during training. Seven perturbations are considered: three synthetic noises, namely Gaussian, Poisson, and uniform noise, and four feature-level approximations of common wearable sEMG artifacts, namely multiplicative gain variation, motion-artifact bursts, baseline drift, and power-line contamination. For each noise type we sweep ten intensity levels from 0.02 to 0.20 and report the mean F1.

Since the input is extracted features rather than raw signals, the four physically motivated artifacts are implemented at the feature level. Multiplicative gain variation is modeled as per-element stochastic scaling to emulate electrode-impedance and gain fluctuation. Motion artifacts are approximated as sample-level transient bursts via Bernoulli-masked heavy-tailed perturbations on a subset of test windows. Baseline drift is a slow-varying trend given by a normalized cumulative random walk modulated by a per-feature direction vector. Power-line contamination is a fixed feature-space offset shared across samples, representing consistent interference after feature extraction.
Table~\ref{tab:noise} shows that the proposed models remain consistently stable across seven perturbations. SDH is tuned only on uniform noise but generalizes to physically motivated artifacts, suggesting that its robustness is not noise-specific.

\begin{figure}[h]
    \centering
    \includegraphics[width=0.5\textwidth]{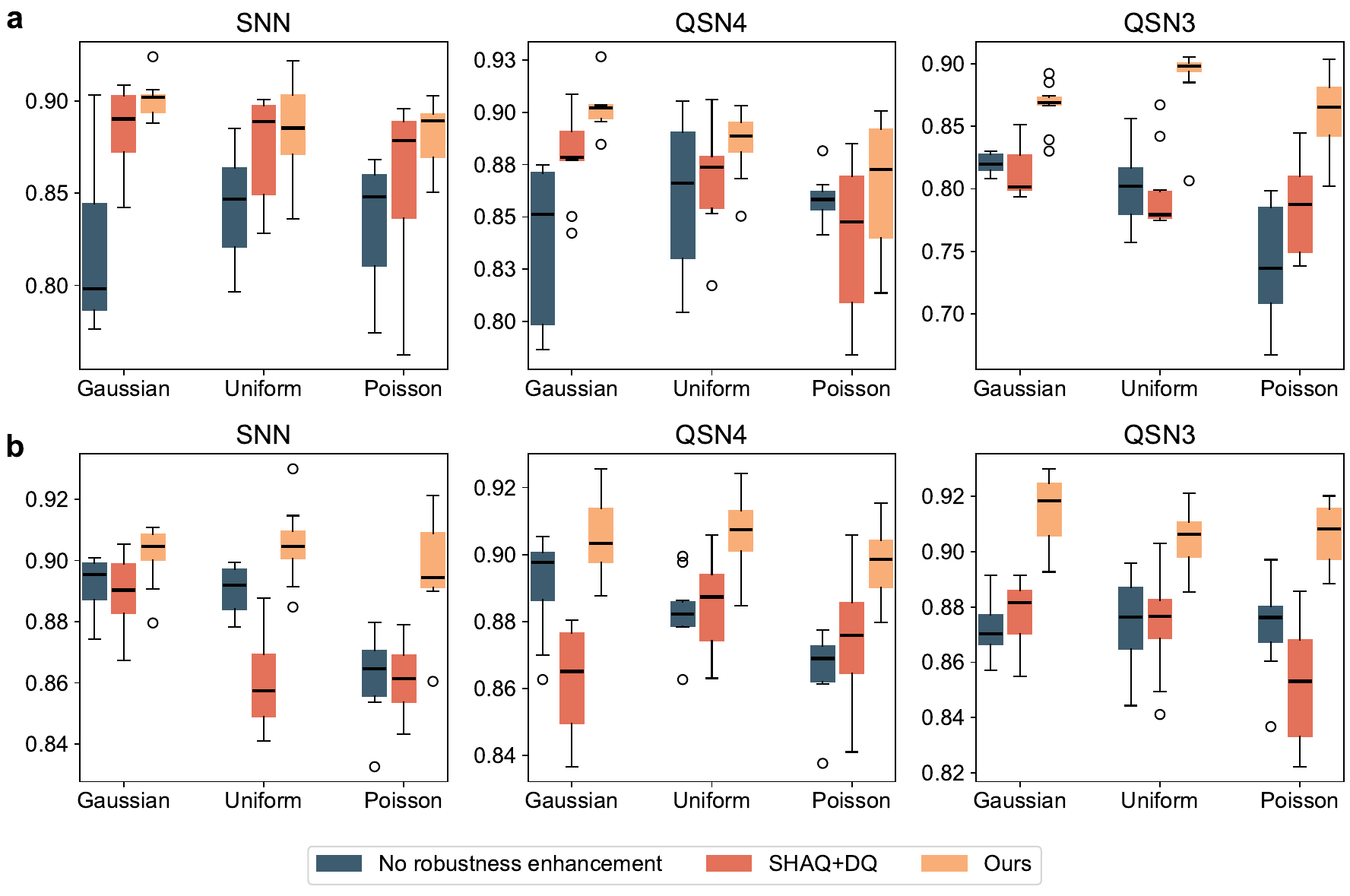}
    
    \caption[]{Ablation study comparing our methods with previous work in terms of F1-score. ``No robustness enhancement'' refers to directly adding noise to the testing data and evaluating using the original model. ``SHAQ+DQ'' consists of three loss terms: binary cross-entropy, $L_{\text{act}}$ for SHAQ, and $L_{\text{orth}}$ for DQ. ``Ours'' represents our proposed composite method, with the coefficients $\lambda$ determined through grid search and kept consistent for the same model and noise. All experiments were performed with the same random seeds. \textbf{a.} Evaluated on the SUE dataset; \textbf{b.} Evaluated on the SPE dataset.}
    \label{fig:noise_aba}
\end{figure}

To further isolate the contribution of SDH, we compared our method against variants trained with no robustness, and with prior methods such as SHAQ and DQ. As shown in Figure~\ref{fig:noise_aba}, SDH outperforms all alternatives across both datasets, achieving superior stability under increasing noise.

\subsection{Energy Consumption}
\label{Energy_consumption}
In this section, we evaluate the energy efficiency of our models. Unlike conventional ANNs, SNNs operate in an event-driven manner, where a simple addition is triggered only by spikes to reduce power consumption. These advantages are further amplified on neuromorphic hardware. Moreover, applying weight quantization further enhances efficiency without sacrificing performance.

To quantify energy savings, we estimate operation counts and energy consumption. In 45nm CMOS technology, a single Multiply-and-Accumulate (MAC) operation consumes 4.6pJ 
, while an Accumulate (AC) operation consumes 0.9pJ~\cite{horowitz20141}. For the baseline MLP, the total computational cost amounts to 21,024 multiplications and 161 additions, yielding an estimated energy consumption of 96.85nJ.

In contrast, for our SNN models, the matrix multiplications in the ANN are replaced by additions, and the number of additions required can be expressed as $N_{\text{ADD}} = T \cdot r \cdot N_{\text{MUL}}$,
where $T$ is the timestep and $r$ is the spike rate.
We measure the spike rate and compute the corresponding energy cost based on low-bit additions in 22nm technology: 0.01626pJ for 4-bit and 0.01167pJ for 3-bit additions. To ensure fair comparison with prior estimates under 45nm technology, we conservatively scale these values by a factor of 2. The average spike rate for our spike-based models is approximately $17.1\%$. This implies that for the \(T=15\) and \(T=6\) models, each spike train requires $2.57\times$ and $1.02\times$ the number of additions, respectively. The resulting energy estimates for each model are shown in Figure~\ref{fig:aba}c and Figure~\ref{fig:aba}f. Our results indicate that, depending on the timestep, SNNs improve energy efficiency by 1.89$\times$--5.13$\times$, while QSNs achieve larger savings of 52.07$\times$--201.77$\times$.

We also validate our approach by deploying the proposed SNNs on the low-power reconfigurable ASIC neuromorphic systems shown in Section~\ref{sec: arc}. With 4-bit weights and time steps $T=15$, the total simulated energy consumption is $19.55 \text{ nJ}$ (accounting for core computation, control, memory, and leakage). To contextualize this, we estimate the operational lifetime of a typical smartwatch with a $4000 \text{ J}$ ($1.1 \text{ Wh}$) battery~\cite{huang2015weardrive} operating at $1 \text{ MHz}$~\cite{chu2022neuromorphic,yan2024reconsidering}. Our method enables approximately $56.83$ hours of continuous operation, demonstrating the superior energy efficiency.

\section{Ablation study}
\label{Ablation}
In this section, we present two ablation studies regarding our choice of features and timesteps.

\subsection{Feature Analysis}
\label{Ablation_Feature}
\begin{figure}[h]
    \centering
    \includegraphics[width=0.45\textwidth]{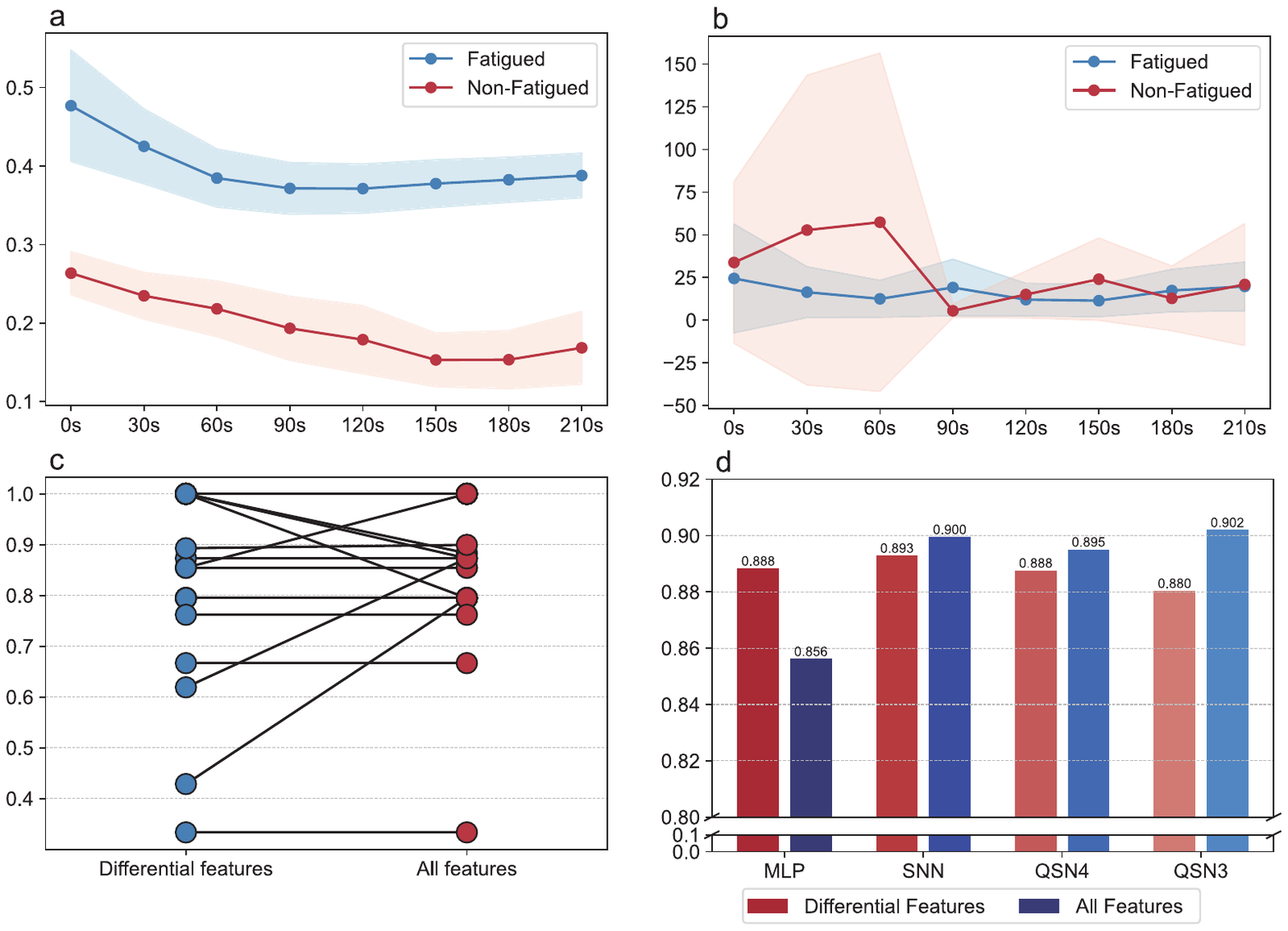}
    \caption{Comparison of differential features and all features. \textbf{a.} Spectral entropy mean values of scaled EMG show significant differences in fatigue and non-fatigue muscle. \textbf{b.} WAMP mean values of scaled EMG show non-significant differences in fatigue and non-fatigue muscle. \textbf{c.} Comparison of F1 score from 30 users. \textbf{d.} Comparison of F1 score by features setting. All shaded areas represent the 95\% confidence interval of the measurements from 30 users.}
    \label{fig:feature}
\end{figure}

We analyze the contribution of different feature subsets. Among the 132 extracted features (see Section~\ref{Preprocessing}), 105 showed significant distribution differences between fatigued and non-fatigued states, indicating clear physiological distinctions (Figure~\ref{fig:feature}a). However, using only these features does not yield optimal accuracy. For instance, WAMP~\cite{ma2024laguerre,sonmezocak2021machine}, though effective for detecting contraction events, struggles to capture gradual fatigue accumulation due to muscle adaptation (Figure~\ref{fig:feature}b).
Interestingly, including the remaining 27 features showing no significant differences further improves performance. As shown in Figure~\ref{fig:feature}c and d, incorporating all 132 features increases the F1 score from 89.3\% to 90.0\% in the full SNN, from 88.8\% to 89.5\% in QSN4, and from 88.0\% to 90.2\% in QSN3. This highlights the complementary value of less discriminative features in enhancing generalization.


\begin{figure}[h]
    \centering
    \includegraphics[width=0.5\textwidth]{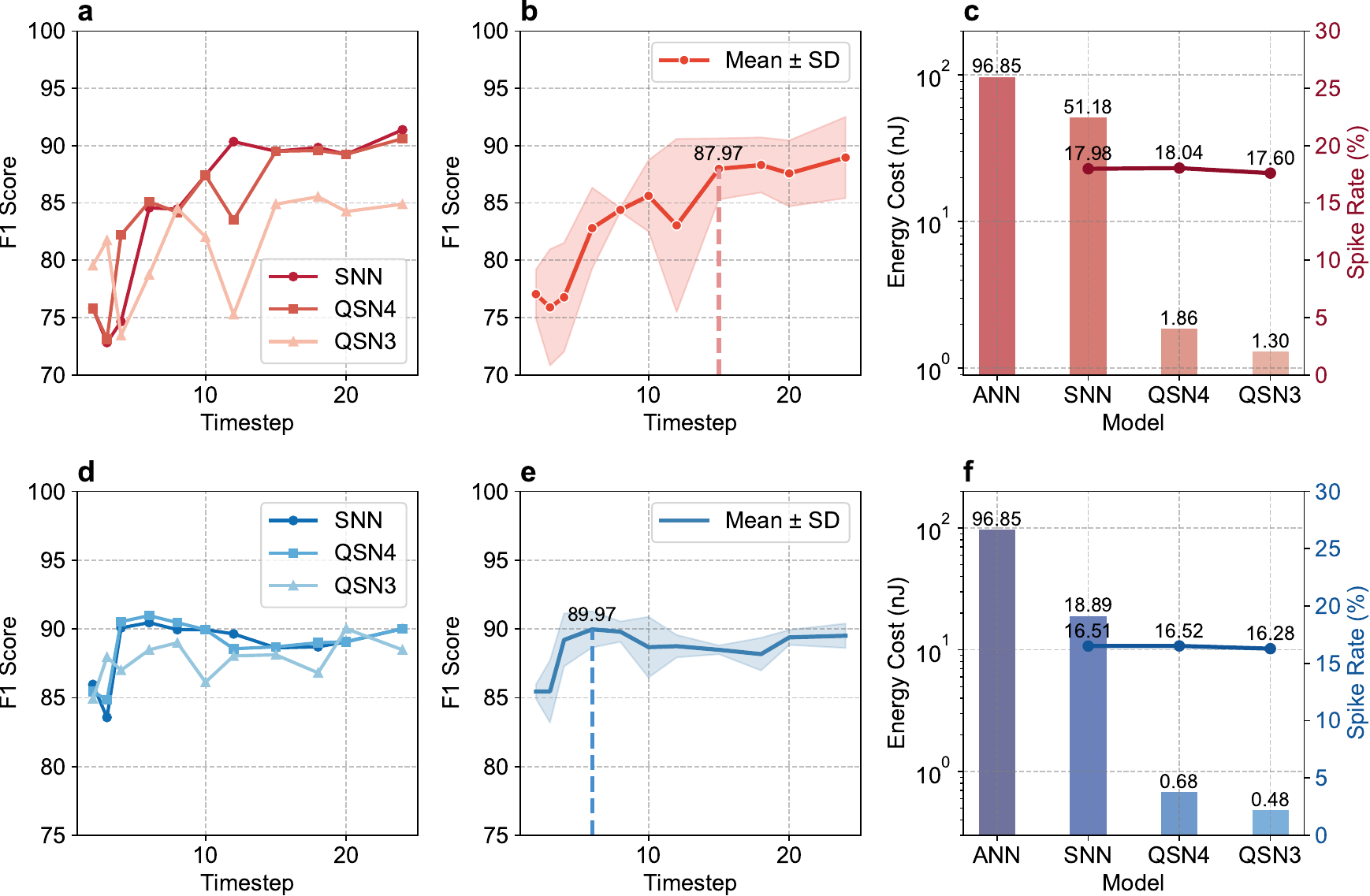}
    \caption[]{
    Ablation study on timestep and energy analysis. \textbf{a, d} Performance variation with timesteps from 2 to 24 for the SUE and SPE datasets, respectively. \textbf{b, e} Averaged F1-scores with standard deviation, highlighting the optimal timesteps (T=15 for SUE and T=6 for SPE). \textbf{c, f} Energy costs comparisons between the baseline MLP and our spike-based models. Spike rates are collected and shown for each model.}
    \label{fig:aba}
\end{figure}

\subsection{Timestep Analysis}
\label{Ablation_Timestep}
A critical factor influencing both the model's performance and energy efficiency is the timestep, denoted as $T$. 
We conducted an ablation study by varying $T$ from 2 to 24 on both datasets. The F1-scores of all spike-based models under different $T$ values are shown in Figure~\ref{fig:aba}.

From Figure~\ref{fig:aba}a, we observe that on SUE dataset, $T=15$ consistently delivers strong performance, achieving a high F1 score while maintaining manageable computational overhead. Increasing $T$ beyond 15 only results in marginal performance gains, while smaller values lead to degraded performance. Consequently, we fix $T=15$ as the default for all models on SUE to ensure strong performance without additional tuning overhead. In contrast, as shown in Figure~\ref{fig:aba}e for the SPE dataset, a timestep of $T=6$ provides sufficient information for effective learning. Larger timesteps in this case tend to lead to overfitting or yield negligible performance improvements.

\section{Conclusion}
\label{Conclusion}

In this work, we propose an energy-efficient sEMG-based muscle fatigue detection approach using SNNs. Through ANN-to-SNN conversion and aggressive 3- and 4-bit weight quantization, we substantially reduce computational costs, achieving up to 200× lower energy consumption than baseline ANNs. Our spiking models maintain performance comparable to full-precision models, enabling real-time deployment on resource-constrained edge devices. We further introduce an advanced training method to improve robustness against environmental noise, which is critical for real-world sEMG acquisition. This work lays the foundation for practical wearable systems for real-time muscle fatigue monitoring.

\bibliographystyle{IEEEtran}
\bibliography{ref}

@misc{
yan2024improving,
title={Improving model robustness against noise with safe haven activations},
author={Zhanglu Yan and Dogukan Yigit Polat and Shida Wang and Kaiwen Tang and Weng-Fai Wong},
year={2024},
url={https://openreview.net/forum?id=PoSq0B0ffE}
}

@article{yan2022low,
  title={Low Latency Conversion of Artificial Neural Network Models to Rate-encoded Spiking Neural Networks},
  author={Yan, Zhanglu and Zhou, Jun and Wong, Weng-Fai},
  journal={arXiv preprint arXiv:2211.08410},
  year={2022}
}

@article{chen2021psychophysiological,
  title={Psychophysiological data-driven multi-feature information fusion and recognition of miner fatigue in high-altitude and cold areas},
  author={Chen, Shoukun and Xu, Kaili and Yao, Xiwen and Zhu, Siyi and Zhang, Bohan and Zhou, Haodong and Guo, Xin and Zhao, Bingfeng},
  journal={Computers in biology and medicine},
  volume={133},
  pages={104413},
  year={2021},
  publisher={Elsevier}
}

@article{puce2021surface,
  title={Surface electromyography spectral parameters for the study of muscle fatigue in swimming},
  author={Puce, Luca and Pallecchi, Ilaria and Marinelli, Lucio and Mori, Laura and Bove, Marco and Diotti, Daniele and Ruggeri, Piero and Faelli, Emanuela and Cotellessa, Filippo and Trompetto, Carlo},
  journal={Frontiers in Sports and Active Living},
  volume={3},
  pages={644765},
  year={2021},
  publisher={Frontiers Media SA}
}

@article{wang2021muscle,
  title={A muscle fatigue classification model based on LSTM and improved wavelet packet threshold},
  author={Wang, Junhong and Sun, Shaoming and Sun, Yining},
  journal={Sensors},
  volume={21},
  number={19},
  pages={6369},
  year={2021},
  publisher={MDPI}
}

@inproceedings{rozaqi2019design,
    title={Design of analog and digital filter of electromyography},
    author={Rozaqi, Latif and Nugroho, Asep and Sanjaya, Kadek Heri and Simbolon, Artha Ivonita},
    booktitle={2019 International Conference on Sustainable Energy Engineering and Application (ICSEEA)},
    pages={186--192},
    year={2019},
    organization={IEEE}
}

@inproceedings{papakostas2019physical,
  title={Physical fatigue detection through EMG wearables and subjective user reports: a machine learning approach towards adaptive rehabilitation},
  author={Papakostas, Michalis and Kanal, Varun and Abujelala, Maher and Tsiakas, Konstantinos and Makedon, Fillia},
  booktitle={Proceedings of the 12th ACM international conference on pervasive technologies related to assistive environments},
  pages={475--481},
  year={2019}
}

@article{lim2024assessment,
  title={Assessment of Self-report, Palpation, and Surface Electromyography Dataset During Isometric Muscle Contraction},
  author={Lim, Jihoon and Lu, Lei and Goonewardena, Kusal and Liu, Jefferson Zhe and Tan, Ying},
  journal={Scientific Data},
  volume={11},
  number={1},
  pages={208},
  year={2024},
  publisher={Nature Publishing Group UK London}
}

@article{ma2024laguerre,
  title={A Laguerre--Volterra network model based on ant colony optimization applied to evaluate EMG-force relationship in the muscle fatigue state},
  author={Ma, Min and Luo, Xi and Xiahou, Shiji and Shan, Xinran},
  journal={Review of Scientific Instruments},
  volume={95},
  number={6},
  year={2024},
  publisher={AIP Publishing}
}

@article{sonmezocak2021machine,
  title={Machine learning and regression analysis for diagnosis of bruxism by using EMG signals of jaw muscles},
  author={Sonmezocak, Temel and Kurt, Serkan},
  journal={Biomedical Signal Processing and Control},
  volume={69},
  pages={102905},
  year={2021},
  publisher={Elsevier}
}

@article{bendtsen1995pressure,
  title={Pressure-controlled palpation: a new technique which increases the reliability of manual palpation},
  author={Bendtsen, L and Jensen, R and Jensen, NK and Olesen, J},
  journal={Cephalalgia},
  volume={15},
  number={3},
  pages={205--210},
  year={1995},
  publisher={SAGE Publications Sage UK: London, England}
}

@article{sun2022application,
  title={Application of surface electromyography in exercise fatigue: a review},
  author={Sun, Jiaqi and Liu, Guangda and Sun, Yubing and Lin, Kai and Zhou, Zijian and Cai, Jing},
  journal={Frontiers in Systems Neuroscience},
  volume={16},
  pages={893275},
  year={2022},
  publisher={Frontiers Media SA}
}

@article{zhang2021mffnet,
  title={MFFNet: Multi-dimensional Feature Fusion Network based on attention mechanism for sEMG analysis to detect muscle fatigue},
  author={Zhang, Yongqing and Chen, Siyu and Cao, Wenpeng and Guo, Peng and Gao, Dongrui and Wang, Manqing and Zhou, Jiliu and Wang, Ting},
  journal={Expert Systems with Applications},
  volume={185},
  pages={115639},
  year={2021},
  publisher={Elsevier}
}

@article{corvini2022estimation,
  title={Estimation of mean and median frequency from synthetic sEMG signals: Effects of different spectral shapes and noise on estimation methods},
  author={Corvini, Giovanni and D'Anna, Carmen and Conforto, Silvia},
  journal={Biomedical Signal Processing and Control},
  volume={73},
  pages={103420},
  year={2022},
  publisher={Elsevier}
}

@article{xu2021advances,
  title={Advances and disturbances in sEMG-based intentions and movements recognition: A review},
  author={Xu, Hao and Xiong, Anbin},
  journal={IEEE Sensors Journal},
  volume={21},
  number={12},
  pages={13019--13028},
  year={2021},
  publisher={IEEE}
}

@article{anwer2024evaluation,
  title={Evaluation of data processing and artifact removal approaches used for physiological signals captured using wearable sensing devices during construction tasks},
  author={Anwer, Shahnawaz and Li, Heng and Antwi-Afari, Maxwell Fordjour and Mirza, Aquil Maud and Rahman, Mohammed Abdul and Mehmood, Imran and Guo, Runhao and Wong, Arnold Yu Lok},
  journal={Journal of Construction Engineering and Management},
  volume={150},
  number={1},
  pages={03123008},
  year={2024},
  publisher={American Society of Civil Engineers}
}

@article{li2022application,
  title={Application of an EMG interference filtering method to dynamic ECGs based on an adaptive wavelet-Wiener filter and adaptive moving average filter},
  author={Li, Yurong and Su, Zhichao and Chen, Kai and Zhang, Wenxuan and Du, Min},
  journal={Biomedical Signal Processing and Control},
  volume={72},
  pages={103344},
  year={2022},
  publisher={Elsevier}
}

@article{wan2017muscle,
  title={Muscle fatigue: general understanding and treatment},
  author={Wan, Jing-jing and Qin, Zhen and Wang, Peng-yuan and Sun, Yang and Liu, Xia},
  journal={Experimental \& molecular medicine},
  volume={49},
  number={10},
  pages={e384--e384},
  year={2017},
  publisher={Nature Publishing Group}
}

@inproceedings{guan2021sports,
  title={Sports fatigue detection based on deep learning},
  author={Guan, Xiaole and Lin, Yanfei and Wang, Qun and Liu, Zhiwen and Liu, Chengyi},
  booktitle={2021 14th International Congress on Image and Signal Processing, BioMedical Engineering and Informatics (CISP-BMEI)},
  pages={1--6},
  year={2021},
  organization={IEEE}
}

@article{al2011review,
  title={A review of non-invasive techniques to detect and predict localised muscle fatigue},
  author={Al-Mulla, Mohamed R and Sepulveda, Francisco and Colley, Martin},
  journal={Sensors},
  volume={11},
  number={4},
  pages={3545--3594},
  year={2011},
  publisher={Molecular Diversity Preservation International (MDPI)}
}

@article{moniri2020real,
  title={Real-time forecasting of sEMG features for trunk muscle fatigue using machine learning},
  author={Moniri, Ahmad and Terracina, Dan and Rodriguez-Manzano, Jesus and Strutton, Paul H and Georgiou, Pantelis},
  journal={IEEE Transactions on Biomedical Engineering},
  volume={68},
  number={2},
  pages={718--727},
  year={2020},
  publisher={IEEE}
}

@article{seshadri2019wearable,
  title={Wearable sensors for monitoring the internal and external workload of the athlete},
  author={Seshadri, Dhruv R and Li, Ryan T and Voos, James E and Rowbottom, James R and Alfes, Celeste M and Zorman, Christian A and Drummond, Colin K},
  journal={NPJ digital medicine},
  volume={2},
  number={1},
  pages={71},
  year={2019},
  publisher={Nature Publishing Group UK London}
}

@article{bangaru2022automated,
  title={Automated and continuous fatigue monitoring in construction workers using forearm EMG and IMU wearable sensors and recurrent neural network},
  author={Bangaru, Srikanth Sagar and Wang, Chao and Aghazadeh, Fereydoun},
  journal={Sensors},
  volume={22},
  number={24},
  pages={9729},
  year={2022},
  publisher={MDPI}
}

@article{zhang2015real,
  title={A real-time, practical sensor fault-tolerant module for robust EMG pattern recognition},
  author={Zhang, Xiaorong and Huang, He},
  journal={Journal of neuroengineering and rehabilitation},
  volume={12},
  pages={1--16},
  year={2015},
  publisher={Springer}
}

@article{eshraghian2023training,
  title={Training spiking neural networks using lessons from deep learning},
  author={Eshraghian, Jason K and Ward, Max and Neftci, Emre O and Wang, Xinxin and Lenz, Gregor and Dwivedi, Girish and Bennamoun, Mohammed and Jeong, Doo Seok and Lu, Wei D},
  journal={Proceedings of the IEEE},
  volume={111},
  number={9},
  pages={1016--1054},
  year={2023},
  publisher={IEEE}
}

@inproceedings{tang2024onespike,
  title={OneSpike: Ultra-low latency spiking neural networks},
  author={Tang, Kaiwen and Yan, Zhanglu and Wong, Weng-Fai},
  booktitle={2024 International Joint Conference on Neural Networks (IJCNN)},
  pages={1--8},
  year={2024},
  organization={IEEE}
}

@article{chu2022neuromorphic,
  title={A neuromorphic processing system with spike-driven SNN processor for wearable ECG classification},
  author={Chu, Haoming and Yan, Yulong and Gan, Leijing and Jia, Hao and Qian, Liyu and Huan, Yuxiang and Zheng, Lirong and Zou, Zhuo},
  journal={IEEE Transactions on Biomedical Circuits and Systems},
  volume={16},
  number={4},
  pages={511--523},
  year={2022},
  publisher={IEEE}
}

@article{yan2023cq,
  title={CQ+ Training: Minimizing Accuracy Loss in Conversion From Convolutional Neural Networks to Spiking Neural Networks},
  author={Yan, Zhanglu and Zhou, Jun and Wong, Weng-Fai},
  journal={IEEE Transactions on Pattern Analysis and Machine Intelligence},
  volume={45},
  number={10},
  pages={11600--11611},
  year={2023},
  publisher={IEEE}
}

@article{lin2019defensive,
  title={Defensive quantization: When efficiency meets robustness},
  author={Lin, Ji and Gan, Chuang and Han, Song},
  journal={arXiv preprint arXiv:1904.08444},
  year={2019}
}

@article{boyer2023reducing,
  title={Reducing noise, artifacts and interference in single-channel EMG signals: A review},
  author={Boyer, Marianne and Bouyer, Laurent and Roy, Jean-S{\'e}bastien and Campeau-Lecours, Alexandre},
  journal={Sensors},
  volume={23},
  number={6},
  pages={2927},
  year={2023},
  publisher={MDPI}
}

@inproceedings{horowitz20141,
  title={1.1 computing's energy problem (and what we can do about it)},
  author={Horowitz, Mark},
  booktitle={2014 IEEE international solid-state circuits conference digest of technical papers (ISSCC)},
  pages={10--14},
  year={2014},
  organization={IEEE}
}

@article{xu2024eescn,
  title={EESCN: A novel spiking neural network method for EEG-based emotion recognition},
  author={Xu, FeiFan and Pan, Deng and Zheng, Haohao and Ouyang, Yu and Jia, Zhe and Zeng, Hong},
  journal={Computer methods and programs in biomedicine},
  volume={243},
  pages={107927},
  year={2024},
  publisher={Elsevier}
}

@inproceedings{fan2024ultra,
  title={An ultra-low power time-domain based SNN processor for ECG classification},
  author={Fan, Haodong and Chang, Liang and Zhou, Junlu and Yang, Xi and Lin, Shuisheng and Zhou, Jun},
  booktitle={2024 IEEE International Symposium on Circuits and Systems (ISCAS)},
  pages={1--5},
  year={2024},
  organization={IEEE}
}

@inproceedings{huang2015weardrive,
  title={WearDrive: Fast and Energy-Efficient Storage for Wearables},
  author={Huang, Jian and Badam, Anirudh and Chandra, Ranveer and Nightingale, Edmund B},
  booktitle={2015 USENIX Annual Technical Conference (USENIX ATC 15)},
  pages={613--625},
  year={2015}
}

@article{yan2024reconsidering,
  title={Reconsidering the energy efficiency of spiking neural networks},
  author={Yan, Zhanglu and Bai, Zhenyu and Wong, Weng-Fai},
  journal={arXiv preprint arXiv:2409.08290},
  year={2024}
}

@inproceedings{chu2023energy,
  title={An Energy-Efficient and Robust SNN Classifier for LC-ADC Sampled ECG Signals},
  author={Chu, Haoming and Yan, Yulong and Zhou, Yiren and Zhu, Zikai and Huan, Yuxiang and Zheng, Lirong and Zou, Zhuo},
  booktitle={2023 8th International Conference on Integrated Circuits and Microsystems (ICICM)},
  pages={502--506},
  year={2023},
  organization={IEEE}
}

@article{li2024hr,
  title={HR-SNN: An end-to-end spiking neural network for four-class classification motor imagery brain-computer interface},
  author={Li, Yulin and Fan, Liangwei and Shen, Hui and Hu, Dewen},
  journal={IEEE Transactions on Cognitive and Developmental Systems},
  year={2024},
  publisher={IEEE}
}

@article{xu2023novel,
  title={A novel event-driven spiking convolutional neural network for electromyography pattern recognition},
  author={Xu, Mengjuan and Chen, Xiang and Sun, Antong and Zhang, Xu and Chen, Xun},
  journal={IEEE Transactions on Biomedical Engineering},
  volume={70},
  number={9},
  pages={2604--2615},
  year={2023},
  publisher={IEEE}
}

@article{guo2024spgesture,
  title={SpGesture: Source-Free Domain-adaptive sEMG-based Gesture Recognition with Jaccard Attentive Spiking Neural Network},
  author={Guo, Weiyu and Sun, Ying and Xu, Yijie and Qiao, Ziyue and Yang, Yongkui and Xiong, Hui},
  journal={arXiv preprint arXiv:2405.14398},
  year={2024}
}

@article{bu2023optimal,
  title={Optimal ANN-SNN conversion for high-accuracy and ultra-low-latency spiking neural networks},
  author={Bu, Tong and Fang, Wei and Ding, Jianhao and Dai, PengLin and Yu, Zhaofei and Huang, Tiejun},
  journal={arXiv preprint arXiv:2303.04347},
  year={2023}
}
\end{document}